\pdfoutput=1

\documentclass[11pt]{article}

\usepackage{authblk}

\usepackage{emnlp2021}

\usepackage{times}
\usepackage{latexsym}

\usepackage[T1]{fontenc}

\usepackage[utf8]{inputenc}

\usepackage{microtype}

\usepackage{booktabs}
\usepackage{graphicx}

\usepackage{nccmath}

\usepackage{arydshln}

\usepackage[ruled,vlined]{algorithm2e}
\usepackage{algorithmic}
\usepackage{booktabs}
\usepackage{csquotes}

\usepackage{multirow}

\usepackage{color}
\definecolor{deepblue}{rgb}{0,0,0.5}
\definecolor{deepred}{rgb}{0.6,0,0}
\definecolor{deepgreen}{rgb}{0,0.5,0}
\definecolor{gael}{RGB}{204, 0, 255}

 \newcommand{\mc}[1]{\textcolor{black}{#1}}
 \newcommand{\ggc}[1]{\textcolor{black}{#1}} \newcommand{\cc}[1]{\textcolor{black}{#1}}
 
\definecolor{gael}{RGB}{0, 0, 0}
\newcommand{\ggf}[1]{\textcolor{black}{#1}}
\newcommand{\llf}[1]{\textcolor{black}{#1}}

\DeclareFixedFont{\ttb}{T1}{txtt}{bx}{n}{9} 
\DeclareFixedFont{\ttm}{T1}{txtt}{m}{n}{9}  

\newcommand\pythoninline[1]{{\pythonstyle\lstinline!#1!}}

\SetKwInput{KwInput}{Input}
\SetKwInput{KwOutput}{Output} 

\usepackage{listings}
\usepackage{color}
\definecolor{deepblue}{rgb}{0,0,0.5}
\definecolor{deepred}{rgb}{0.6,0,0}
\definecolor{deepgreen}{rgb}{0,0.5,0}

\newcommand\pythonstyle{\lstset{
language=Python,
basicstyle=\ttm,
otherkeywords={self},             
keywordstyle=\ttb\color{deepblue},
emph={MyClass,__init__},          
emphstyle=\ttb\color{deepred},    
stringstyle=\color{deepgreen},
frame=tb,                         
showstringspaces=false            %
}}

\lstnewenvironment{python}[1][]
{
\pythonstyle
\lstset{#1}
}
{}

\usepackage{amsmath}

\title{\ggc{ Few-Shot Emotion Recognition in Conversation with Sequential Prototypical Networks }}

\author[1,2]{Gaël Guibon}
\author[1]{Matthieu Labeau}
\author[2]{Hélène Flamein}
\author[2]{Luce Lefeuvre}
\author[1]{Chloé Clavel}
\affil[1]{LTCI, Télécom-Paris, Institut Polytechnique de Paris}
\affil[2]{Direction Innovation \& Recherche SNCF}
\affil[ ]{\textit {\{gael.guibon,matthieu.labeau,chloe.clavel\}@telecom-paris.fr}}
\affil[ ]{\textit {\{ext.gael.guibon,helene.flamein,luce.lefeuvre\}@sncf.fr}}

\begin{document}
\maketitle
\begin{abstract}

Several recent studies on dyadic human-human interactions have been done on conversations without specific business objectives. However, many companies might benefit from studies dedicated to more precise environments such as after sales services or customer satisfaction surveys. In this work, we place ourselves in the scope of a live chat customer service in which we want to detect emotions and their evolution in the conversation flow. This context leads to multiple challenges that range from exploiting restricted, small and mostly unlabeled datasets to finding and adapting methods for such context.
We tackle these challenges by using Few-Shot Learning while making the hypothesis it can serve conversational emotion classification for different languages and sparse labels. We contribute by proposing a variation of Prototypical Networks for sequence labeling in conversation that we name ProtoSeq. We test this method on two datasets with different languages: daily conversations in English and customer service chat conversations in French. When applied to emotion classification in conversations, our method proved to be competitive even when compared to other ones. 
\ggf{The code for ProtoSeq is available at \url{https://github.com/gguibon/ProtoSeq}.}

\end{abstract}

\section{Introduction}
\label{sec:intro}

\mc{There has been a recent surge in research focusing on analyzing dyadic human to human interactions}. Many of these studies \cite{poria2017context,zadeh2018memory,zadeh2018multi,majumder_dialoguernn_2019} focus on emotion recognition in conversations (ERC) taking into account multiple data modalities. Moreover, 
\mc{most of the progress made in ERC has been done without factoring in constraints corresponding to specific but prominent industrial applications, like customer service}. This is partly due to studies focusing on using artificial datasets \cite{li_dailydialog_2017,busso2008iemocap} made of mock-up conversations to facilitate result replication and comparison. \cc{A few existing studies address customer service applications \cite{mundra2017fine,yom2018customer,maslowski2017wild} and show the difficulties to deal with such in-the-wild and domain-specific data.  } 

In this work,
\mc{we focus on data from}
a live chat support in which we want to detect emotions and their evolution in the conversational flow. \mc{This setting corresponds to}
a human dyadic conversation, albeit with a specific business-related objective. We make the hypothesis that the emotion flows of the visitor and \ggf{the} operator will
\mc{bring information on} the quality of the service and help operators better assist customers. 
\ggf{This hypothesis is close to relevant studies on the importance of emotions and empathy in dyadic call center conversations \cite{alam_computational_2017,alam2018annotating}.}
\mc{This specific setting leads to multiple challenges: indeed, it is difficult and costly to label this kind of data --- and even then, these exchanges are very sparse in emotions, most of the labels associated with utterances being neutral. To maximize data efficiency, we use Few-Shot Learning (FSL), and adapt a popular approach to our highly unbalanced data. }
By setting up this approach in an episodic fashion \cite{ravi2016optimization}, we join studies on ERC and studies on FSL to tackle this industrial use-case.

We contribute by proposing a variant to Prototypical Networks \cite{snell2017prototypical} dedicated to ERC \mc{on data produced by company services, framing it as} a sequence labeling task. 
We modify the original 
\mc{model} by \mc{allowing it to consider the whole conversational context when making predictions, through a sequential context encoder and the use of Conditional Random Fields (CRF) on top of the model.}
We test our method on two datasets, in two different languages. \mc{The first one, made of daily conversations in English, allows us to compare ourselves to previous methods,  while the second one, made of private data from a live chat customer service, allows us to conduct a performance analysis in our target setting.}
We also present the latter 
dataset, along with \mc{its annotation process.}

This paper is organized as follows. First, we sum up the related work on textual ERC and FSL in conversations (Section~\ref{sec:relatedwork}). Then we present the datasets along with the emotional annotation scheme and the annotation campaign set up for the customer service live chats dataset (Section~\ref{sec:data}). We continue by thoroughly presenting the Sequential Prototypical Networks (Section~\ref{sec:methodo}) before looking at the achieved results on both datasets (Section~\ref{sec:results}). Finally, we present the limitations of such a system (Section~\ref{sec:limitations}) and conclude (Section~\ref{sec:conclusion}).

\section{Related Work}
\label{sec:relatedwork}

\paragraph{Emotion Recognition in Conversations}
\ggc{In recent years, \mc{the widening scope of emotion detection tasks led to the rise of another sub-topic:} 
detecting emotions in conversations. This research topic, commonly referred to as ERC, gained \mc{popularity} when \citet{poria2017context} first applied recurrent neural networks (RNN) \cite{jordan1997serial} to multi-modal emotion recognition in conversations. This led to many improvements \cite{zadeh2018memory,zadeh2018multi,hazarika2018conversational,majumder_dialoguernn_2019}. Among those, \citet{majumder_dialoguernn_2019} used \ggf{3 Gated Recurrent Units (GRU)} \cite{cho2014learning} units, one for each context representation target (speaker, utterance, emotion). \mc{Studies on ERC applied to text followed, mainly built on} an artificial conversation dataset named DailyDialog \cite{li_dailydialog_2017}. \cite{zhong_knowledge-enriched_2019} incorporated a knowledge base into the network using context-aware attention and hierarchical self-attention using Transformers \cite{vaswani2017attention}. \citet{ghosal-etal-2019-dialoguegcn} uses graph neural networks to deal with context propagation limitations. These approaches in ERC consider the conversational context surrounding the current utterance; on the other hand, some recent studies consider it as a sequence and tackled \mc{ERC} through a sequence labeling task \cite{wang-etal-2020-contextualized}. }
 \ggc{We 
\mc{follow this last approach}
and consider the ERC task as a sequence labeling task. However, 
\mc{these supervised approaches are difficult to use, as it is hard} to find a sufficient amount of conversations labeled with emotions. Hence, in this paper, we approach ERC as a few-shot learning problem.}

\paragraph{Few-Shot Learning}
\ggc{FSL \cite{miller2000learning,fei2006one,lake_lakeetal2015science-startoffewshotpdf_2015} is suitable to tackle this data limitation. It aims at generalizing faster, leading to a lower dependency on data quantity. It is mainly set up through episodic composition \cite{ravi2016optimization} which recreates the few-shot learning setting by working with small training episodes. Several learning methods are based on metric-learning:  Siamese Networks, which share some weights, are used to learn a metric between examples \cite{koch_siamese_2015}. Matching Networks \cite{vinyals_matching_2017} use \ggc{the training examples to find the weighted nearest neighbors} \cite{vinyals_matching_2017}. Prototypical Networks \citet{snell2017prototypical} consider \ggc{averaged class representations from the training examples and a cosine distance to compare the elements to these class representations.} Relation networks replace the Euclidean by the deep neural network which aims at training a distance metric \cite{sung2018learning}. }

\noindent \ggc{In this work, we consider approaches based on Prototypical Networks.  As \mc{\citet{pmlr-v130-al-shedivat21a} recently showed it, such approaches are the most efficient when working with a low} amount of training samples. \mc{Many variants have been proposed,} on different tasks and topics such as relation classification in text \cite{gao2019hybrid,hui2020few,ren2020two}, sentiment classification in Amazon comments \cite{bao_few-shot_2020}, named entity recognition \cite{fritzler2019few,hou2020few,perl-etal-2020-low,safranchik2020weakly}, or even speech classification in conversation \cite{koluguri2020meta}. }
\ggc{This surge of interest on applying few-shot learning to these topics can be attributed to specific datasets, such as Few-Rel \cite{han_fewrel_2018} for relation classification. While ERC is mainly considered \mc{in a fully supervised learning setting, we intend to view it} as a few-shot learning sequence labeling class.} 
\ggc{In this paper, we propose the first few-shot learning approach on ERC using sequence labeling through adapting Prototypical Networks. We compare our method to the original Prototypical Networks \cite{snell2017prototypical} and to a variant dedicated to named entity recognition 
\cite{fritzler2019few} that is easily applicable to our task.}

\section{Data}
\label{sec:data}

\mc{To be able to both study the behavior of our model in its targeted industrial use-case, and allow performance comparison with baselines, we will work with two very different corpora: our proprietary live chat customer service dataset, and DailyDialog \cite{li_dailydialog_2017}.}
In both datasets, messages are labeled \mc{with emotions} while considering the context of the conversation. However, they vary considerably in their topics and lexical fields: ordinary matters for DailyDialog and \mc{railway related customer service for the live chats}. They also vary \mc{in the 
assumptions they make about the speakers 
}: \mc{while the topics discussed in DailyDialog imply a sense of proximity, the live chat customer service}  
involve complete strangers with pre-existing emotional states (\textit{e.g.} the visitor is already stressed due to a refund issue). \ggf{Both datasets' statistics can be found in Table \ref{tab:datacounts}.}

\subsection{DailyDialog} 
DailyDialog is a dyadic conversation dataset in English \mc{whose purpose is to represent casual, everyday interactions}
between people, \mc{in order to facilitate training and sharing of dialog systems}. The 
\mc{exchanges} in DailyDialog are \mc{artificial} conversations \mc{which are neither dedicated to a specific topic nor task-oriented: they mainly deal with}
relationships, everyday life, and work.
\mc{Each utterance corresponds to a speaker turn, and} is labeled with one of 7 labels: the 6 basic emotions (anger, disgust, fear, joy, sadness, and surprise) and  "no\_emotion" denoting the absence of one. The "no\_emotion" label represents 80\% of the corpus, leading to a very unbalanced dataset with an average length of 8 messages per conversation and a maximum of 35 messages. For this dataset, the inter annotator agreement achieved 78.9\%.
We choose DailyDialog 
\mc{for comparison and reproducibility purposes, as it is often used for ERC.}
In this work, we use the train/val/test splits provided by \cite{zhong_knowledge-enriched_2019}.

\subsection{
\mc{Live chat customer service}}
Our 
\mc{primary}
objective is to detect emotions in conversations from a \mc{customer service live} chat involving a visitor (\textit{i.e.} the customer looking for help) and an operator (\textit{i.e.} an employee being there to assist the visitor and better satisfy him). The corpus is written in French and is made of 5,000 conversations from which we annotate a subset of 1,500 conversations, leading to a total of 20,754 messages. The average message length is higher than DailyDialog, with \ggf{15.14} messages per conversation. 
\ggf{We do not have a way to identify real speaker turns. Indeed, a speaker turn is not necessarily the sequence of contiguous segments corresponding to a same speaker because there could be a time delay between two messages of a same speaker, indicating that the speaker is changing the topic. Because all our messages have a very short time difference we prefer not to automatically infer speaker turns and consider the message as the unit of analysis. This means the conversation context is a sequence of messages instead of a sequence of speaker turns which could have contained one or more messages artificially glued together.}

\begin{table}[htbp]
    \centering
    \color{gael}
    \begin{tabular}{c|c|c}
        Dataset & DD & Chat \\
        \toprule
        Language & English & French \\
        Type & Artificial & Customer Service\\
        Max Msg/Conv & 35 & 84 \\
        Avg Msg/Conv & 8 & 13 \\
        Labels & 7 & 11 \\
        Labels for eval & 6 & 9 \\
        Nb. Conv. & 13,118 & 1,500 \\
    \end{tabular}
    \caption{\ggf{Statistics for both datasets DailyDialog (DD) and Live Chat Customer Service (chat).}}
    \label{tab:datacounts}
\end{table}

\begin{figure}[htbp]
    \centering
    \includegraphics[width=0.45\textwidth]{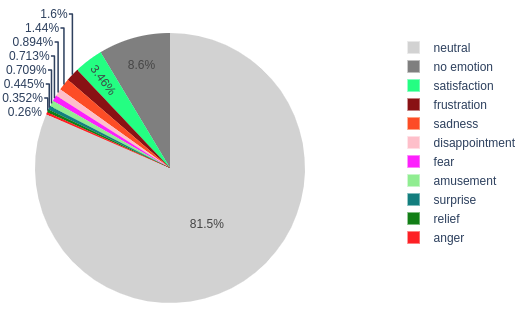}
    \caption{Emotion Distribution in Live Chat Customer Service}
    \label{fig:emotion_count_pie}
\end{figure}

\noindent \mc{Two annotators were involved in the process, which unrolled as follows: first, each message is labeled with an emotion. Once all the messages in a conversation have been assigned an emotion label, the conversation is labeled with a visitor satisfaction score (ranging from -3 to 3), and the status of the customer request ("solved", "test\_required", "out of scope", or "aborted").} 
\cc{After a preliminary study of the corpus, we identify} 10 emotion labels as relevant in this corpus: neutral, surprise, amusement, satisfaction\footnote{\cc{It is interesting to notice here that in the current application setting, "joy" label has been replaced by "satisfaction", because it is more suited to the customer relationship context \cite{danesi2010impact}.}}, relief, fear-anxiety-stress, sadness, disappointment, anger, and frustration. 
\ggf{Compared to \cite{chowdhury2016predicting}, we consider the satisfaction at the conversation level and we are more precise with not only positive, neutral, and negative \llf{levels}, but also \llf{with} 4 additional intermediate levels (from -3 to +3 included). We have also a higher number of emotions, with 10 emotions instead of 4, with more precise emotions such as relief for instance.} 
\ggf{In our customer service interface, some alerts are automatically prompted for specific actions such as “user x left the chat” or “operator sent a link”. \llf{We call these “alerts”, and they are labeled as “no\_emotion”. The “neutral” label means that the emotional content of the message, written by a human, has been considered as neutral by the annotator.} 
Figure \ref{fig:emotion_count_pie} illustrates the distribution of emotion labels in the Live Chat Customer Service dataset. We can see that the neutral label is the most frequent by a large margin.}

The Cohen's $\kappa$ scores obtained on the 3 label types correspond to substantial agreement at the message level and moderate agreement at the conversation level \cite{landis1977application}. \ggf{$\kappa$-score is given for 3 label types: 1) the emotions at the message level ($\kappa=0.65$); 2) the visitor's satisfaction at the conversation level ($\kappa=0.45$); and 3) the request’s status at the conversation level ($\kappa=0.46$).} 
Similarly to DailyDialog, the "neutral" label represents 81.5\% of the corpus, resulting in another very unbalanced dataset in terms of emotions, as rendered obvious by Figure \ref{fig:emotion_count_pie}. Excluding this label gives a slightly more balanced label set, as the satisfaction represents 44.9\% of the other emotions, and the "frustration" 20.8\%. 

\noindent To tackle our hypothesis that the conversational emotion flow can define the overall visitor satisfaction, we calculate the Pearson correlation between the emotions \cc{at the message level }and \cc{the global }satisfaction of the visitor \cc{at the conversation level}. These scores show the more extreme the emotion, the greater the correlation with the satisfaction score is\footnote{\ggc{See appendix \ref{sec:correlations}}}. 

\begin{table}[!h]
    \centering
    \color{gael}
    \begin{tabular}{c|c}
        Emotion & $\kappa$-score \\
        \toprule
        Amusement & 0.1115 \\
        Anger & 0.1608 \\
        Disappointment & 0.1609 \\
        Frustration & 0.1193 \\
        Neutral & 0.3187 \\
        Fear & 0.1111 \\
        Satisfaction & 0.2068 \\
        Relief & 0.1429 \\
        Surprise & 0.1885 \\
        Sadness & 0.2860 \\
        \hline
        Global & 0.6499 \\
        Global w/o Neutral and no\_emotion & 0.3885 \\     
        \bottomrule
    \end{tabular}
    \caption{\ggf{By-category agreement scores for emotions in Live Chat Customer Service}}
    \label{tab:by-category-k-scores}
\end{table}



\section{Methodology}
\label{sec:methodo}

\mc{Formally, our dataset $\mathcal{D}$ is comprised of conversations $(C_1,C_2,\dots,C_{|D|})$, which are in turn made of utterances: $C_i= (u_1,u_2,\dots,u_{|C_i|})$. To each of these utterances is associated an emotion label, giving a sequence of labels by conversation: $Y_i = (y_1, y_2, \dots, y_{|C_i|})$. Finally, an utterance is a sequence of words, $u_j = (w^j_1, w^j_2, \dots, w^j_{|u_j|})$
.}

\subsection{\mc{Episodic learning}}

\ggc{We use the episodic approach \cite{ravi2016optimization}, which simulates a context where only a few examples per class are available during training and the model must adapt during testing. This approach perfectly fits into our need for FSL.} The episodic composition is defined by setting the number of classes (ways) $N_{\mathcal{C}}$, the number of examples per class $N_S$ (shots) and the number of elements to label $N_Q$ (queries). In our experiments with DailyDialog, the task is 5-shot 7-way 10-query, and when using our customer service chats, the number of classes changes, making it a 5-shot 11-way 10-query. In the context of \mc{sequential} ERC, this means that for each episode we train the model on 5 conversations (\textit{i.e.} sequences to label) per emotion and \mc{apply it to} 10 conversations per emotion. \mc{We identify a sequence as belonging to the target class set if at least one message is labeled with the target class in the sequence. This means that the number of example messages in each support set $S_k$ of class $k$ can vary (with a minimum of $N_S$ elements), while the number of sequences is fixed}. 

\subsection{ProtoSeq: Prototypical Networks for \ggc{Emotion} Sequence Labeling}

\mc{In order to apply \ggf{FSL} to ERC, we choose to base our model on Prototypical Networks~\cite{snell2017prototypical}, which} create prototypes from the average of the embeddings of the words forming the \mc{utterance}. Our proposed model, ProtoSeq, \mc{builds on this by factoring in conversational context and performing sequence labeling, thus allowing the use of both input and output dependencies when applying \ggf{FSL} to ERC.} 
\mc{ProtoSeq is divided into four main components, applied at each consecutive level of granularity of the data.}

\begin{figure}[htbp]
    \centering
    \includegraphics[width=.45\textwidth]{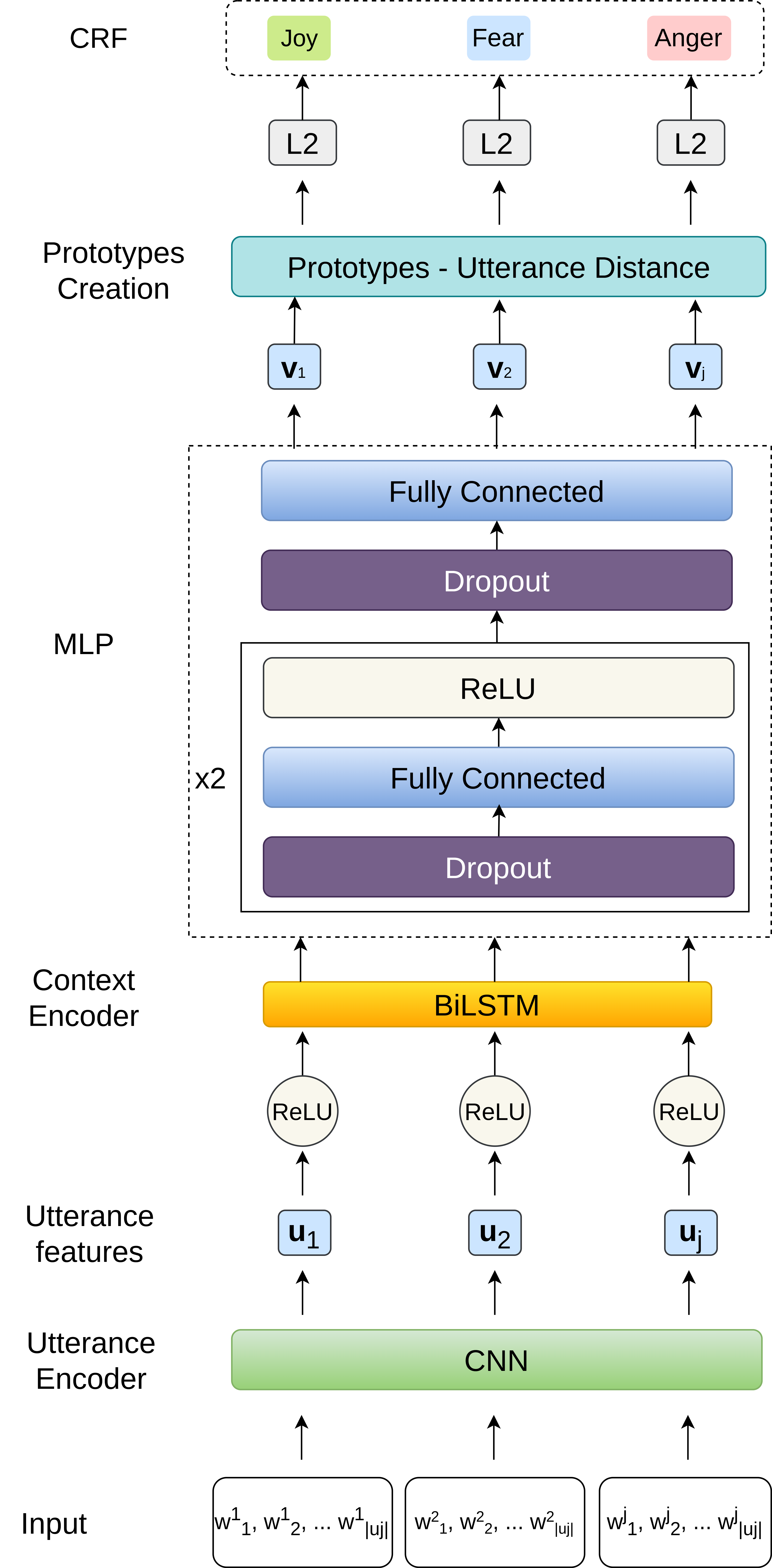}
    \caption{ProtoSeq Global View}
    \label{fig:protoseq}
\end{figure}

\paragraph{Utterance Encoder:}
\mc{Similar to the 
encoder of the Prototypical Networks, our utterance encoder $f_u$ reduces the utterance $u_i$ to only one vector:
$$ \mathbf{u}_j = f_u(w^j_1, w^j_2, \dots, w^j_{|u_j|}) $$
} 
\mc{
The architecture of our encoder is based on the
Convolutional Neural Network (CNN) 
described by ~\cite{kim2014convolutional}
, which makes} tokens through different convolution filters and merges the representation through max-over-time pooling. 

\paragraph{Context Encoder:}
After applying a non-linear activation (ReLU), we use a Bi-directional LSTM layer \ggf{(BiLSTM)} \cite{huang2015bidirectional}, \mc{to integrate}
contextual information from the conversation, 
\mc{thus following}
the trend initiated by \cite{poria2017context} in ERC to use a recurrent context encoder. \mc{We obtain contextual utterance representations $\mathbf{v}_j$:
$$ \mathbf{v}_j = f_c([\mathbf{u}_k]_{k=1}^{j-1}, \mathbf{u}_j, [\mathbf{u}_k]_{k=j+1}^{|C_i|})$$}
As we work in a few-shot learning setting, we try not to over-complexify our model, hence we do not add \mc{a transformer-based} global context encoder \cite{wang-etal-2020-contextualized} 
on top of the BiLSTM. 

\paragraph{Prototypes Creation:}
\mc{We feed the output of the context encoder}
to a multi-layer perceptron made of 2 fully connected layers with dropout and ReLU. The resulting representations are then used to create prototypes $\mathbf{c}$: \mc{for the class $k$, 
$$\mathbf{c}_k \gets \frac{1}{N_{\mathcal{C}}} \sum\limits_{(u_j, y_j) \text{ with } y_j = k} MLP(\mathbf{v}_j)$$ 
where $N_{\mathcal{C}}$ is the number of classes, \ggf{and MLP refers to Multi-Layer Perceptron.}}

\paragraph{Sequence Prediction:}
We compute the \mc{euclidean} distance from the
contextual representation of the utterance to each class prototype. 
\mc{The predicted label $\hat{y}_j$ to each utterance $u_j$ is the class corresponding to its closest prototype:}
$$\begin{aligned} \hat{y}_j \gets \arg\min_{k \in \mathcal{C}} d( MLP(\mathbf{v}_j), \mathbf{c}_{k}) \end{aligned}$$

We allow our model to consider dependencies between the labels, 
we add a final CRF layer on top of label prediction, the emission scores being the euclidean distances for each utterance. 
\mc{Overall, our model is a variation of the traditional BiLSM-CRF model, based on prototypical networks.}

\subsection{Experimental protocol}

We follow the setting used by \cite{bao_few-shot_2020} by considering a training epoch as a set of 100 random episodes from the training set, and applying a validation step made of 100 random episodes from the validation set after each epoch. We test our model using 1,000 random episodes from the test set. The maximum number of epochs is set to 1,000, but when the F1-micro score does not improve for 100 consecutive epochs, we stop the training and reload the best model's weights. We use the Adam \cite{kingma2017adam,loshchilov2019decoupled} optimizer to train the model while maximizing the log-likelihood loss of the correct emotion sequences in the query set $Q_k$

$$\mathcal{L}=\sum_{C \in Q_k} \sum_{j=1}^{|C|} \log (p({\hat{y}_j} \mid u_j, C)) $$
  
During inference, we apply the Viterbi algorithm \mc{to output the best-scoring sequence of labels. We do not cut on either utterance or conversation length}. 
To obtain an initial token representation, we use pre-trained FastText \cite{bojanowski2017enriching} embeddings from Wiki News\footnote{\url{https://dl.fbaipublicfiles.com/fasttext/vectors-english/wiki-news-300d-1M.vec.zip}} for English (DailyDialog), and from Common Crawls\footnote{\url{https://dl.fbaipublicfiles.com/fasttext/vectors-crawl/cc.fr.300.bin.gz}} for French (customer service live chats). Both sets of embeddings are of dimension $300$ and both datasets are tokenized with NLTK\footnote{\url{https://www.nltk.org/}}. 
We choose our hyper-parameters using a very targeted grid search for the learning rate (\ggc{set to $1e3$ for all the experiments}) and manual tuning for the other parameters. \mc{In the following, we experiment with several variants of our model, each having  dedicated hyper-parameters.}
\begin{itemize}
    \item \textbf{ProtoSeq}: We use hyper-parameters from \citet{kim2014convolutional} for the CNN: 50 filters with windows 3 different sizes (3, 4 and 5). We use one BiLSTM layer with $150$ hidden units in order to fit to the $300$ dimensions of the inputs considering the two directions.
    \item \textbf{ProtoSeq-CNN}: A lighter version of our model, without the BiLSTM context-encoder. The CNN configuration follows the same parameters from \citet{kim2014convolutional}.
    \item \textbf{ProtoSeq-Tr}: A ProtoSeq with a 2-layers Transformer\cc{-based} \ggc{utterance} encoder with 4 attention heads and a hidden size of $300$. The global dropout is set to $0.2$ while the position encoder dropout is set to $0.1$. 
    \item \textbf{ProtoSeq-AVG}: \ggc{A ProtoSeq where the utterance encoder is just an average of the token representations. 
    However, it should be noted that the averaging process excludes the padding elements in the utterances. }
\end{itemize}

\section{Results}
\label{sec:results}

\begin{table*}[h!]
    \centering
    
\begin{tabular}{clll}
    \toprule
    Model  &  F1 (weighted) & MCC & F1 (micro) \\ 
    \midrule
    \multicolumn{4}{c}{\bf Supervised Learning} \\
    \midrule
    cLSTM  & & & 0.4990 \\
    CNN \cite{kim2014convolutional} & & & 0.4934 \\
    CNN+cLSTM \cite{poria2017context} & & & 0.5184 \\
    BERT BASE \cite{devlin2018bert} & & & 0.5312 \\ 
    DialogueRNN \cite{majumder_dialoguernn_2019}  & & & 0.5164 \\
    KET \cite{zhong_knowledge-enriched_2019} & & &  0.5337 \\
    CESTa \cite{wang-etal-2020-contextualized} & &  & \bf 0.6312 \\
    \midrule 

    \multicolumn{4}{c}{\bf Few-Shot Learning}\\
    \midrule
    Proto \cite{snell2017prototypical} &  0.2377 \tiny $\pm  0.0136$ &  0.3448 \tiny $\pm  0.0105$ &  0.2141 \tiny $\pm  0.0141$ \\
    \ggc{WarmProto-CRF} \cite{fritzler2019few} &  \ggc{0.2384} \tiny $\pm  0.0383$ &  \ggc{0.3403} \tiny $\pm  0.0365$ &  \ggc{0.2607} \tiny $\pm  0.0381$ \\
    \midrule
    ProtoSeq-AVG & 0.1312 \tiny $\pm  0.0201$ &  0.2622 \tiny $\pm  0.0225$ &  0.1643 \tiny $\pm  0.0258$ \\
    ProtoSeq-Tr  &  0.1694 \tiny $\pm  0.0293$ &  0.3329 \tiny $\pm  0.0241$ &  0.2557 \tiny $\pm  0.0317$\\
    ProtoSeq-CNN &  0.2244 \tiny $\pm  0.0359$ &  0.3494 \tiny $\pm  0.0182$ &  0.2560 \tiny $\pm  0.0275$ \\
    ProtoSeq &  0.3522 \tiny $\pm  0.0302$ &  0.3922 \tiny $\pm  0.0233$ &  \bf 0.3181 \tiny $\pm  0.0276$ \\
    \bottomrule
    \end{tabular}
    \caption{Sequence labeling on DailyDialog splits \cite{zhong_knowledge-enriched_2019} (seq size = 35). Top section shows supervised learning results reported from related work, bottom section presents our results using few-shot learning (7 way 5 shot 10 query). \ggc{MCC = multi-class Matthews Correlation Coefficient (MCC). $\pm$ = test episodes variance..}} 
    \label{tab:res_convseq_dailydialog_tiny}
\end{table*}

\begin{table*}[h!]
    \centering
    \color{gael}
\begin{tabular}{clll}
    \toprule
    Model &   F1 (weighted) & MCC & F1 (micro) \\ 
    \midrule
    \multicolumn{4}{c}{\bf Supervised Learning} \\
    \midrule
    KET \cite{zhong_knowledge-enriched_2019} & & & \bf 0.4143 \\
    \midrule 

    \multicolumn{4}{c}{\bf Few-Shot Learning}\\
    \midrule
    Proto \cite{snell2017prototypical} &  0.1749 \tiny $\pm  0.0481$ &  0.3333 \tiny $\pm  0.0133$ &  0.1228 \tiny $\pm  0.0194$ \\
    \ggc{WarmProto-CRF} \cite{fritzler2019few} &  \ggc{0.1556} \tiny $\pm  0.0522$ &  \ggc{0.7212} \tiny $\pm  0.0220$ &  \ggc{0.1918} \tiny $\pm  0.0601$ \\
    \midrule
    ProtoSeq-AVG  &  0.1297 \tiny $\pm  0.0246$ &  0.7163 \tiny $\pm  0.0215$ &  0.1582 \tiny $\pm  0.0251$ \\
    ProtoSeq-Tr & 0.1774 \tiny $\pm  0.0285$ &  0.6695 \tiny $\pm  0.0163$ &  0.2208 \tiny $\pm  0.0371$ \\
    ProtoSeq-CNN  &  0.1197 \tiny $\pm  0.0198$ &  0.7336 \tiny $\pm  0.0135$ &  0.1581 \tiny $\pm  0.0180$ \\
    ProtoSeq  &  0.3022 \tiny $\pm  0.0256$ &  0.6396 \tiny $\pm  0.0222$ & \bf 0.2668 \tiny $\pm  0.0270$ \\
    \bottomrule
    \end{tabular}
    \caption{Few-shot learning results on Customer Service Live Chats (seq size = 18): 11-way 5-shot 10-query (padding \& trim). \ggc{MCC = multi-class Matthews Correlation Coefficient (MCC). $\pm$ = test episodes variance.}  }
    \label{tab:res_convseq_ouitchat_tiny}
\end{table*}

\cc{Tables \ref{tab:res_convseq_dailydialog_tiny} and \ref{tab:res_convseq_ouitchat_tiny} show the performance of the model using the micro F1-score. We use the protocol usually followed by the literature and do not take into account the majority class "no\_emotion" as it represents 80\% of the DailyDialog corpus. } 
This allows performance comparison with related work on ERC through supervised learning. We do the same for the Live Chat Customer Service corpus by ignoring the "neutral" label. 

\paragraph{Comparison to supervised learning}
DailyDialog is used to compare our FSL approach with recent supervised learning results on ERC. As expected, our best FSL model, ProtoSeq, yields lesser performance than supervised approaches. \cc{The latter presuppose the availability of a sufficiently large amount of annotated data} and their performance thus represents the upper bound of the expected results. 
More precisely, \mc{we focus on the difference between ProtoSeq with a state-of-the-art supervised model, CESTa \cite{wang-etal-2020-contextualized}, which is computation-heavy. \ggc{Indeed, CESTa is a contextualized emotion sequence tagging model which considers the fusion of \ggf{a combination of a transformer and BiLSTM as the global context encoder} with a recurrent individual context encoder before feeding a CRF layer. CESTa achieves 63\% in micro F1-score in a fully supervised learning approach.}  ProtoSeq, much lighter, achieves a 31\% micro F1 score, demonstrating the potential of FSL for sequence labeling when available data is scarce, \ggc{especially when many supervised approaches obtained F1-scores around 50\%.}} 
\ggf{While using the Live Chat Customer Service dataset, we only change the initial embeddings from English to French, and apply the two best models according to \ref{tab:res_convseq_dailydialog_tiny}: CESTa and KET \cite{zhong_knowledge-enriched_2019}. The CESTa implementation yielded inconclusive results\footnote{\ggf{We present in our code an implementation of CESTa following the paper's descriptions. On our dataset, it only labeled the two majority classes 'no\_emotion' and 'neutral', leading to a null F1 (micro).}}, this is why we present the KET results on our specific corpus in Table \ref{tab:res_convseq_ouitchat_tiny}. KET relies on ConceptNet \cite{speer2017conceptnet}, a multilingual knowledge base. Thus, we only switch from GloVe embeddings \cite{pennington2014glove} to French FastText ones in order to ensure comparison with our ProtoSeq model. As expected, performance is lower on the Live Chat Customer Service corpus.}

\paragraph{Few-shot learning baselines}
We consider two baselines. We apply the original Prototypical Networks \cite{snell2017prototypical}, only retrieving the labels using the euclidean distance to class prototypes. We also apply the WarmProto-CRF \cite{fritzler2019few} which is a variant of Prototypical Networks designed for sequence labeling by integrating CRFs. We implement it without including the bias they created for the O label in the BIO sequence labeling task. This method uses a BiLSTM utterance encoder to further compute the prototypes with the euclidean distance.

\paragraph{Few-shot learning on DailyDialog}
Table \ref{tab:res_convseq_dailydialog_tiny} shows FSL results in the bottom section. All these models are trained in an episodic fashion, with the same episode constitution (5-shot 7-way 10-query). 
We can see the micro F1-score is really low with only 16.43\%. By considering a ProtoSeq only using an utterance encoder based on CNN (ProtoSeq-CNN) or an utterance encoder based on a 2-layers 4-heads Transformer (ProtoSeq-Tr) we can see the score improve. The addition of the BiLSTM context encoder really enables the model to capture more information: 
these variants show the importance of integrating a context encoder in the model. 

\paragraph{Few-shot learning on Customer Service Live Chats}
We also apply this approach on the Customer Service Live Chats\ggf{, further motivated by the high annotation cost and the fact that supervised approaches on clean data such as DailyDialog did not achieve an acceptable score for this use case (starting from 70 \% in micro F1 score). Besides, new conversations with evolving contents (e.g., due to the evolution of company services) are created everyday. As a consequence, it would render the ideally annotated training corpus obsolete at some point.} This \ggf{FSL} prediction leads to lesser scores, but with the same hierarchy among variants. ProtoSeq, using a BiLSTM context encoder, yields again the best scores. \mc{The higher number of classes (with 11 classes including 9 emotions versus 7 classes including 6 emotions) may explain the overall lower numbers we observe here, compared to those we obtain on DailyDialog.} 

\paragraph{Artificial versus Real Data}
DailyDialog is an artificial corpus which follows standard, idealized conversations. We can see that ERC performance is quite sensible to the conversation length, which seems to confirm conclusions drawn in recent literature  \cite{wang-etal-2020-contextualized}. Customer Service Live Chats being real use-case data, their length varies a lot, ranging from 2 to 85 messages (where conversations from DailyDialog go from 2 to 35 messages). However, ERC also seems to be impacted by the utterance textual content, as our data contains a lot of spelling mistakes, shortcuts, or slurs. 
\mc{More importantly, the visitor may often use several small messages rather than only one to transmit information; this flow may be interrupted by a message from the operator,}
making it impossible to detect the whole set of messages as an utterance.
This is specific to online instant conversations where speakers do not necessarily wait for the complete message to be written or sent by the addressee. By contrast, DailyDialog is made of clean and perfect exchanges, where one waits for the other to send the answer. Here is an example with the following clean conversation subset from DailyDialog. 
\begin{quote}
\textbf{A}: Does your family have a record of your ancestors? \\
\textbf{B}: Sure. My mom has been working on our family tree for years.
\end{quote}
This conversation would often be represented as follows in real data from instant chat:
\begin{quote}
\textbf{Operator}: Did you make the simulation using the promo code?\\
\textbf{Visitor}: I did it 5 minutes ago\\
\textbf{Operator}: Ok, you have to wait 30min\\
\textbf{Visitor}: but as said before, I didn't finished the "simulation" because I had to pay a 10€ ticket even th\\
\textbf{Visitor}: ....even though the right one is 11.5€\\
\textbf{Operator}: And the code will be available again\\
\end{quote}
Moreover, specific lexical fields, \mc{relevant to the customer service being provided, can also make the task more difficult for the model.} 

\paragraph{Quantifying the impact of the CRF layer}

\ggc{Our model benefits from the addition of a final CRF layer to compute the best possible \mc{output} sequence. 
This allows the model to generalize faster and to achieve a higher score despite the few examples. However, the prediction stability lowers, as 
the standard deviation across episodes 
\mc{shows} in Table \ref{tab:res_nocrf}. The downgrade in performance while omitting the CRF layer may be due to the label dependency it emphasizes. Indeed, without the CRF, label dependency can only be inferred from the BiLSTM context encoder. The CRF layer accentuates in-episode label dependency by allowing the prediction to be further adapted to the conversation context for each query conversation.}

\begin{table}[htbp]
    \centering
    \small
\begin{tabular}{cll}
    \toprule
    Model & DailyDialog & CSChats \\ 
    \midrule
    ProtoSeq-noCRF & 0.2156 \tiny $\pm  0.0105$ & 0.1351 \tiny $\pm  0.0144$\\
    ProtoSeq & 0.3181 \tiny $\pm  0.0276$ & 0.2668 \tiny $\pm  0.0270$ \\
    \bottomrule
    \end{tabular}
    \caption{Micro F1-scores without and with the final CRF layer. CSChats = Customer service live chat }
    \label{tab:res_nocrf}
\end{table}

\paragraph{Emotion Predictions}

\ggc{Tables \ref{tab:report_dd} and \ref{tab:report_oui} show additional information from ProtoSeq's performance on each label. These tables present averaged scores from all episodes' query sets. We can see the predictions differ a lot depending on the target label. When applied to DailyDialog, the model has no difficulty in detecting the absence of emotion. This is to be expected as this label mainly represents the conversations. However, the prediction scores for emotion labels are imbalanced, with recall scores higher than precision on both datasets.} 

\ggc{On DailyDialog, the anger and the sadness labels really hinders the overall prediction. However, on the Customer Service Live Chats, Table \ref{tab:report_oui} shows really poor prediction for the disappointment (translated from the French "déception" label) and fear labels. Actually, in this dataset the precision seems to be the main issue with only the frustration and satisfaction labels being somewhat correctly labeled.} \ggc{Given the model and the task, the detailed results obtained on both datasets show that performance score may benefit from the usage of macro F1-score along with the micro F1-score. Indeed, be it DailyDialog or Customer Service Live Chats, the multi-class prediction of sequence tagging is really sparse, and thus leads to imbalanced prediction, even while using an episodic strategy.}

\ggc{Moreover, the gap between results on DailyDialog and the ones on the Customer Service Live Chats confirms the necessity for the ERC-related studies to focus on real conversation datasets whenever it is possible.}

\begin{table}[htbp]
\centering
\small
\begin{tabular}{lrrr}
\toprule
{} &  precision &  recall &  f1-score \\
\midrule
no emotion   &       0.98 &    0.91 &      0.94 \\
anger        &       0.24 &    0.38 &      0.30 \\
disgust      &       0.12 &    0.29 &      0.17 \\
fear         &       0.58 &    0.55 &      0.57 \\
happiness    &       0.39 &    0.63 &      0.49 \\
sadness      &       0.07 &    0.21 &      0.11 \\
surprise     &       0.17 &    0.37 &      0.24 \\
\bottomrule
\end{tabular}
\caption{Additional results on DailyDialog with our ProtoSeq prediction. }
\label{tab:report_dd}
\end{table}

\begin{table}[htbp]
    \centering
    \small
    \begin{tabular}{lrrr}
\toprule
{} &  precision &  recall &  f1-score \\
\midrule
no emotion   &       1.00 &    1.00 &      1.00 \\
surprise     &       0.06 &    0.10 &      0.07 \\
amusement    &       0.12 &    0.54 &      0.20 \\
satisfaction &       0.47 &    0.60 &      0.53 \\
relief       &       0.21 &    0.23 &      0.22 \\
neutral      &       0.92 &    0.79 &      0.85 \\
fear*        &       0.02 &    0.01 &      0.01 \\
sadness      &       0.08 &    0.18 &      0.11 \\
disappointment&       0.03 &    0.07 &      0.04 \\
anger        &       0.02 &    0.40 &      0.03 \\
frustration  &       0.45 &    0.43 &      0.44 \\
\bottomrule
\end{tabular}
    \caption{Additional results on customer service live chats with our ProtoSeq prediction. We define the "fear" label as "fear/anxiety/stress". "no emotion" is only used for automatic chat prompts.}
    \label{tab:report_oui}
\end{table}

\section{Limitations}
\label{sec:limitations}

While the ProtoSeq model seems to be suitable for FSL in ERC, it still has inherent limitations \mc{related to its architecture}. ProtoSeq uses a CRF as its final layer, leading to a sequence labeling optimizer that does not take the order into account. While this yields better performance, it does not guarantee that the order information retrieved from the context encoder is wisely used, especially since \mc{we use the euclidean distances to class prototypes as emission scores for the sequence labeling.
An ordered-prediction approach may allow the model to better assist operators in real-time during their decision process.} 

Another limitation of \mc{our model is that it may be difficult to adapt to changes in the context in which customer service is provided. Indeed, the type of service or the plaftorm used may lead to lexical field changes or very different emotional states for the incoming visitors.}

\section{Conclusion}
\label{sec:conclusion}

\ggc{In this paper, we presented the first study on emotion recognition in conversations using few-shot learning. We proposed a variant of Prototypical Networks taking into account the emotion recognition as a sequence labeling task while allowing fast convergence. When compared to other prototypical networks for sequence labeling in few-shot, our model obtained higher scores on both DailyDialog and Customer Service Live chats. }\ggc{Through this work, we showed that few-shot learning is possible for this task even though it is still difficult to achieve the same performance as supervised learning approaches.} \cc{This study also shows the challenges that remain when tackling in-the-wild data collected in the context of a real application. } 

\cc{Future work will be dedicated to the improvement of the current few-shot ERC approach by adding unlabeled elements in the support set and by investigating the addition of external business knowledge to such an approach.}

\section*{Acknowledgements}
This project has received funding from SNCF, the French National Research Agency’s grant ANR-17-MAOI and the DSAIDIS chair at Télécom-Paris.

\bibliography{anthology,custom}
\bibliographystyle{acl_natbib}

\appendix
\cleardoublepage

\section{Correlation scores}
\label{sec:correlations}

\begin{figure}[h!]
    \centering
    \includegraphics[width=.99\textwidth]{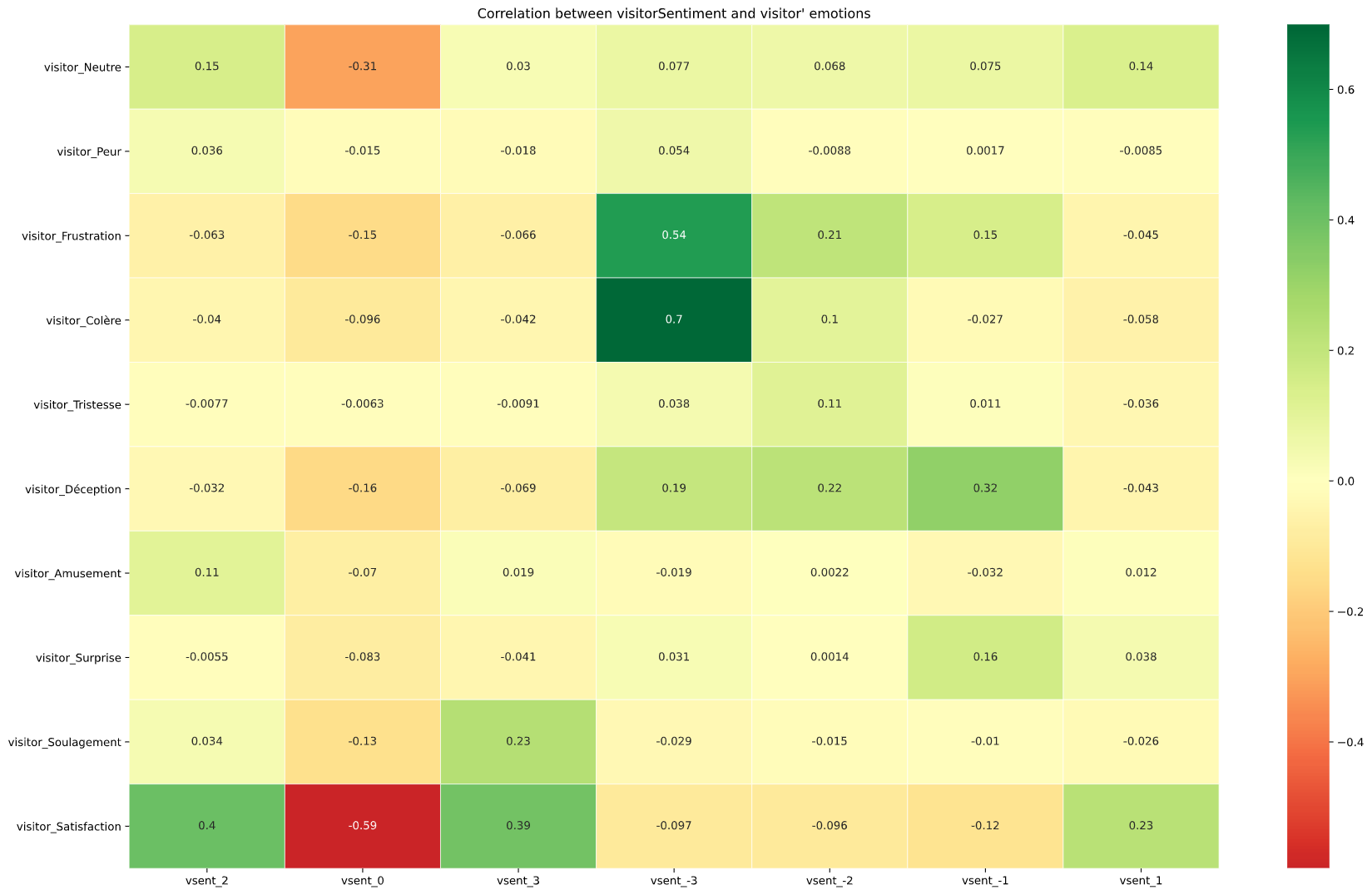}
    \caption{\ggf{Pearson correlation scores between the visitor's overall satisfaction score in the conversation (vsent) and the presence of specific emotions in the messages’ emotion flow (visitor\_\lbrack emotion\rbrack).}}

    \label{fig:corr}
\end{figure}
Figure \ref{fig:corr} presents the Pearson correlation scores between visitor's emotions and satisfaction for the Customer Service Live Chats. While emotions are labeled for each utterance in conversation, satisfaction is a global label for the whole conversation. 
\ggf{This Figure shows the correlation scores are higher when the emotion is 
\llf{extreme within a given polarity.} 
For instance, anger is greatly correlated to a negative satisfaction score (vsent -3) than fear or disappointment, while “Satisfaction” is more correlated to a positive overall satisfaction score (vsent +3) than “Amusement” or “Relief” are to intermediate satisfaction scores (vsent\_1 or vsent\_2).}

\end{document}